\documentclass[11pt]{article}
\usepackage{coling2020}
\usepackage{times}
\usepackage{url}
\usepackage{latexsym}
\usepackage{graphicx}
\usepackage{multirow}
\usepackage{multicol}
\usepackage{makecell}
\usepackage{array}
\usepackage{xcolor}
\usepackage{footnote}
\usepackage{threeparttable}
\usepackage[skip=0.25\baselineskip]{caption}
\usepackage{enumitem}
\usepackage{amssymb}
\usepackage[symbol]{footmisc}

\renewcommand{\thefootnote}{\fnsymbol{footnote}}

\newcommand{\revise}[1]{{\color{red}#1}}

\colingfinalcopy % Uncomment this line for the final submission

\title{An Industry Evaluation of Embedding-based Entity Alignment}
\author{
Ziheng Zhang$^{1}$\footnotemark[1] ,
Jiaoyan Chen$^{2*}$,
Xi Chen$^{1*}$\footnotemark[2] , \\
\textbf{Hualuo Liu$^{1}$,
Yuejia Xiang$^{1}$,
Bo Liu$^{1}$,
Yefeng Zheng$^{1}$} \\
  $^{1}$Tencent Jarvis Lab, Shenzhen, China \\
  $^{2}$Department of Computer Science, University of Oxford, UK \\
  {\small \tt \{zihengzhang,jasonxchen\}@tencent.com, jiaoyan.chen@cs.ox.ac.uk} \\
  {\small \tt lhl18@mails.jlu.edu.cn, \{yuejiaxiang,raymanliu,yefengzheng\}@tencent.com}
}

\date{}

\begin{document}

\maketitle

\footnotetext[1]{The first three authors contributed equally.}
\footnotetext[2]{Xi Chen is the corresponding author.}

\begin{abstract}
Embedding-based entity alignment has been widely investigated in recent years, but most proposed methods still rely on an ideal supervised learning setting with a large number of unbiased seed mappings for training and validation, which significantly limits their usage. In this study, we evaluate those state-of-the-art methods in an industrial context, where the impact of seed mappings with different sizes and different biases is explored. Besides the popular benchmarks from DBpedia and Wikidata, we contribute and evaluate a new industrial benchmark that is extracted from two heterogeneous knowledge graphs (KGs) under deployment for medical applications. The experimental results enable the analysis of the advantages and disadvantages of these alignment methods and the further discussion of suitable strategies for their industrial deployment.
\end{abstract}

\renewcommand{\thefootnote}{\arabic{footnote}}

\section{Introduction}
\label{sec:introduction}
% Structure
% Paragraph 1: Introduce the background of KG alignment and embedding-based methods
% Paragraph 2: Introduce the importance of seeding alignments (training samples) -- most effective methods are supervised or semi-supervised; emphasis the sample shortage challenge in real world application; their is no empirical study on those embedding-based methods in the context of sample shortage.
% Paragraph 3: This study investigates the performance of those embedding-based methods under different settings of sample shortage, and discussed effective sampling strategies for training embedding-based entity alignment models.

Knowledge graphs (KGs), such as DBpedia \cite{auer2007dbpedia}, Wikidata \cite{vrandevcic2014wikidata} and YAGO \cite{suchanek2007yago} are playing an increasingly important role in various applications such as question answering and search engines.
The construction of KGs usually includes several components, such as Named Entity Recognition (NER) \cite{li2018survey}, Relation Extraction (RE) \cite{zhang2019long}, and Knowledge Correction \cite{chen2020correcting}.
However, the content of an individual KG is often incomplete, leading to a limited knowledge coverage especially in supporting applications of a specific domain \cite{farber2018linked,demartini2019implicit}.
One widely adopted solution is to merge multiple KGs (e.g., an enterprise KG with fine-grained knowledge of a specific domain and a general-purpose KG with an extensive coverage) with the assistance of an alignment system which discovers cross-KG mappings of entities, relations, and classes \cite{otero2015ontology,yan2016survey}.

%
%motivated many knowledge-based applications, but KGs are usually incomplete and complementary to each other. 
Embedding-based \textit{entity alignment} has recently attracted more attention due to the popularity of KGs with big data (i.e. a large number of facts) such as Wikidata. 
Traditional alignment systems such as PARIS \cite{suchanek2011paris} and LogMap \cite{jimenez2011logmap}, which usually reply on lexical matching and semantic reasoning (e.g., for checking the violation of relation domain and range), are believed to be weak in utilizing the contextual semantics especially the graph structure of such large KGs.
To address this problem, some novel embedding-based methods have been proposed with the employment of different KG embedding methods such as TransE \cite{bordes2013translating} and Graph Neural Networks (GNNs) \cite{scarselli2008graph} as well as some algorithms from active learning \cite{berrendorf_active_2020}, multi-view learning \cite{zhang_multi-view_2019} and so forth.

%\textbf{Entity alignment} is the technique designed to integrate heterogeneous knowledge among different KGs, and aims to find the entities from different KGs referring to the same real-world object. 
%Each entity, or real-world object,
% is organized into triples of \textit{(head\_entity, relation, tail\_entity)} and 
%can be embedded into low-dimensional vectors with semantics retained using translating embeddings like TransE \cite{bordes_translating_nodate}. 
%The embedding-based entity alignment has attracted much more attention because of the rapid growth in the development of KG embedding approaches.

We find all these embedding-based entity alignment methods rely upon \textit{seed mappings} for supervision or semi-supervision in training.
They are usually evaluated by benchmarks extracted from DBpedia, Wikidata and YAGO, all of which are constructed from the same source, namely Wikipedia.
These methods typically build their models with $30\%$ (or even higher) of all the ground-truth mappings, and the training and validation sets are randomly extracted, sharing the same distribution as the test set.

\begin{figure}
    \centering
    \includegraphics[width=0.45\linewidth]{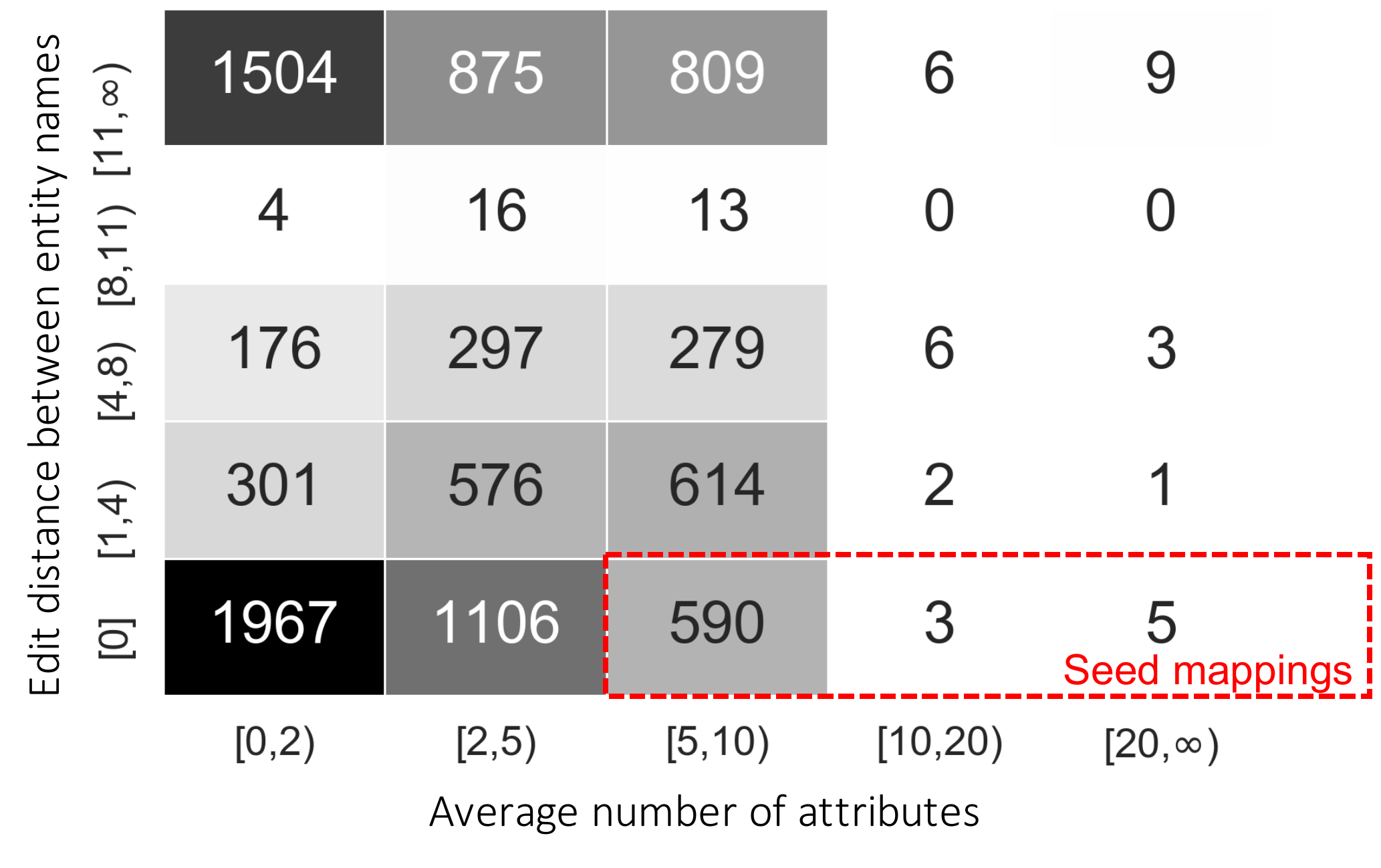}
    \caption{Distribution of mappings of two sampled medical KGs. The horizontal axis denotes the average number of attributes and the vertical axis denotes the edit distance between entity names.}
    \label{fig:heatmap}
\end{figure}

In industrial applications, however, such seed mappings require not only expertise but also much human labour for annotation, especially when the two large KGs come from totally different sources.
Even though a small number of seed mappings can be annotated, they are usually biased in comparison with the remaining for prediction with respect to entity name, attribute, graph structure and so on.
Figure~\ref{fig:heatmap} shows the distribution of all the mappings of two sampled medical KGs from Tencent Technology (cf. Section \ref{sec:ib}
for more details), with two dimensions -- the similarity between names of mapping entities and the average attribute number of mapping entities.
When we directly invited experts or utilized downstream applications to annotate mappings, the annotated mappings, which could act as the seed mappings for training, usually lie in the bottom right area (seen in the red block in Figure~\ref{fig:heatmap}) with high name similarity and large attribute number.
%In our industry application of entity alignment of two medical KGs, we found a significant bias between the seed mappings and the other mappings.
%We firstly hired some annotators to label the aligned entities, \revise{but the annotated seed mappings failed the entity alignment approaches even after several labelling rounds.}
%Figures~\ref{fig:heatmap} depicts the distribution of entity pairs from our industry dataset in terms of the averaged number of attributes and edit distances between entity names. Specifically, larger edit distances indicate lower string similarity of the entity pair, and because some of the entities in the industry dataset are garbled, some edit distances cannot be calculated and are categorized as ``diff" in the heatmap.
%The biased seed mappings are enclosed with a red block in Figure~\ref{fig:heatmap}, and the biases can be described from three aspects:
Thus, we believe that the seed mappings should have the following characteristics to make the evaluation of these supervised methods more practical.
Firstly, the seed mappings should take a small proportion of all the mappings, such as $3\%$ that is far smaller than previous experimental settings.
Secondly, the seed mappings should be biased towards the remaining mappings with respect to the entity name similarity, the average attribute number, or both.
Such biases are ignored in the current evaluation.

In this work, we systematically evaluate four state-of-the-art embedding-based KG alignment methods in an industrial context.
The experiment is conducted with one open benchmark from DBpedia and Wikidata, one industry benchmark from two enterprise medical KGs with heterogeneous contents, and a series of seed mappings with different sizes, name biases and attribute biases.
The performance analysis considers all the testing mappings as well as different splits of them for fine-grained observations.
These methods are also compared with the traditional system PARIS.
To the best of our knowledge, this is the first work to evaluate and analyse the embedding-based entity alignment methods from an industry perspective.
We find that these methods heavily rely on an ideal supervised learning setting and suffer from a dramatic performance drop when being tested in an industrial context.
Based on these results, we can further discuss the possibility to deploy them for real-world applications as well as suitable sampling strategies.
The new benchmark and seed mappings can also benefit the research community for future studies, which are publicly available at \url{https://github.com/ZihengZZH/industry-eval-EA}.

%Therefore, to conduct such a study and gain insights into the impact of seeding alignments on the performance of embedding-based approaches, we present a number of controllable experimental settings, evaluate the state-of-the-art approaches under these settings, and discuss effective sampling strategies for the selection of seeding alignments. 
% Specifically, we investigate four factors of seeding alignments on public datasets, including 1) entity names, 2) degree distribution, 3) seed quantity and 4) properties of each entity. 
%Specifically, we investigate two factors of seeding alignments with few-shot learning, including 1) entity names and 2) attributes of each entity.
%To the best of our knowledge, this work is the first thorough empirical study on the impact of seeding alignments on embedding-based cross-KG entity alignment. 

%Our contributions are summarized as follows.
%\begin{itemize}
%    \setlength\itemsep{-0.1em}
%    \item We experimentally analyse some state-of-the-art approaches for embedding-based entity alignment against the industry settings with limited number of seeding alignments available.
%    \item We investigate two different forms of bias, names and attributes, in the seeding alignments and their impact on the learning of entity embeddings.
%    \item We propose effective seeds sampling strategies that outperform random sampling strategies or achieve comparative performance with less seeding alignments.
%\end{itemize}

\section{Preliminaries and Related Work}

\subsection{Embedding-based Entity Alignment}
\label{sec:eea}
% Introduce the embedding-based entity alignment methods, especially those used for evaluation. Why these methods are selected? Two reasons to consider: representative, state-of-the-art performance. We also need to consider those relying on a small number of seeding alignments.
% model choice: BootEA / MultiKE / RDGCN / RSN4EA
% Introduce those empirical analysis of embedding-based entity alignment; what's the difference between these empirical studies and our study? OpenEA

Most of the existing embedding based entity alignment methods conform to the following three-step paradigm: 
% embedding
\textit{(i)} embedding the entities into a vector space by either a translation based method such as TransE \cite{bordes2013translating} or Graph Neural Networks (GNNs) \cite{scarselli2008graph} which recursively aggregate the embeddings of the neighbouring entities and relations;
% mapping
\textit{(ii)} mapping the entity embeddings in the space of one KG to the space of another KG by learning a transformation matrix, sharing embeddings of the aligned entities, or swapping the aligned entities in the associated triples;
% searching
\textit{(iii)} searching an entity's counterpart in another KG by calculating the distance in the embedding space using metrics such as the cosine similarity.
It is worth noting that the role of the seed mappings mainly lies in the second step, aligning the embeddings of two KGs.

%Most existing approaches represent entities in different KGs as embeddings, and try to find the entity alignment by measuring the similarity, such as cosine similarity, between these embeddings \cite{sun2018bootea} \cite{zhang_multi-view_2019} \cite{wu_relation-aware_2019} \cite{guo2018recurrent}. Specifically, to model the entity embeddings, they employ translational models based on the relation triples, or graph neural networks (GNNs) by recursively aggregating the embeddings of the neighbouring entities.

Specifically, we evaluate four methods, namely \textbf{BootEA} \cite{sun2018bootea}, \textbf{MultiKE} \cite{zhang_multi-view_2019}, \textbf{RDGCN} \cite{wu_relation-aware_2019} and \textbf{RSN4EA} \cite{guo2018recurrent}.
On the one hand, they have achieved the state-of-the-art performance in the ideal supervised learning setting, according to their own evaluation and the benchmarking study \cite{sun_benchmarking_2020};
on the other hand, they are representative to different techniques that are widely used in the literature.
The four methods are introduced as follows.

%our experimental approaches because of their state-of-the-art performance and their utilization of different forms of information into entity embeddings.
%We choose \textbf{BootEA} \cite{sun2018bootea}, \textbf{MultiKE} \cite{zhang_multi-view_2019}, \textbf{RDGCN} \cite{wu_relation-aware_2019}, and \textbf{RSN4EA} \cite{guo2018recurrent} as our experimental approaches because of their state-of-the-art performance and their utilization of different forms of information into entity embeddings.

\textbf{BootEA} is a semi-supervised approach, which adopts translation-based models for embedding and iteratively trains a classifier by bootstrapping. In each iteration, new likely mappings are labelled by the classifier and those causing no conflict are added for training in the following iteration.

\textbf{MultiKE} utilizes multi-view learning to encode different semantics into the prediction model. Specifically, three views are developed for entity names, entity attributes, and the graph structure respectively.

\textbf{RDGCN} applies a GCN variant, Dual-Primal GCN \cite{monti2018dual} to utilize the relation information in KG embedding. It can better utilize the graph structure than those translation-based embedding methods, especially in dealing with the triangular structures.
%incorporates relation information via attentive interactions between KGs and further captures neighbouring structures to learn entity embeddings.

\textbf{RSN4EA} firstly generates biased random walks (long paths) of both KGs as sequences and then learns the embeddings by a sequential model named Recurrent Skipping Network. The seed mappings here are used to generate cross-KG walks, thus exploring correlations between cross-KG entities.

\subsection{Seed Mappings}
% Introduce those works about seeding alignments: 
% 1) methods dealing with seeding alignments e.g., active learning; how do they solve the sample shortage problem?
% 2) works evaluating the role of seeding alignments. What is the shortcomings of these evaluation works? Incomplete?

As far as we know, the current embedding-based entity alignment methods mostly rely on the seed mappings, whose roles are introduces in Section~\ref{sec:eea}, for supervised or semi-supervised learning.
Specially, we can consider some heuristic rules with, for example, string and attribute matching to generate the seed mappings, as done by the method IMUSE \cite{he2019unsupervised}, but the impact of the seed mappings is similar and the study of such impact also benefit the distant supervision methods.

%
%Seeding alignments are significantly important for most existing approaches because very few studies for entity alignment can be categorized into unsupervised learning, in which IMUSE \cite{he2019unsupervised}, for instance, relies on the heuristically produced seeding alignments via comparing the string similarity of attributes. 

In addition, although some semi-supervised approaches such as BootEA \cite{sun2018bootea} and SEA \cite{pei2019semi} are less dependent on the seed mappings, their performance, when trained on a small set of seed mappings, may vary from data to data and be impacted by the bias of the seed mappings.  
%indeed eliminate the need for massive seeding alignments \cite{sun2018bootea} \cite{iptranse}, different initial seeding alignments could lead to different alignment results.

In the own evaluation of these methods and the recent benchmark study \cite{sun_benchmarking_2020}, $20\%$ and $10\%$ of all the ground truth mappings are used for training and validation respectively, and more importantly, they are randomly selected, thus maintaining the same distribution as the testing mappings.
This violates the real-world scenarios in the industry, where annotating seed mappings is costly and the annotated ones are usually biased, as discussed in Section \ref{sec:introduction}.
%For most previous experiments, the seeding alignments typically accounts for 20\% of the entire dataset, while the number of seeding alignments is quite limited in the industry applications (around 3\%). 
%In addition, 
Actually, there are relatively few studies that investigate the seed mappings and those investigated only consider the proportion of the seeding mappings. 
In \newcite{sun2018bootea} and \newcite{wu_relation-aware_2019}, the proposed methods are evaluated with the proportion of the seed mapping for training varying from $10\%$ to $40\%$. 
However, the minimum proportion still leads to a very large number (e.g., 1.5K) of seed mappings in aligning two big KGs.

%and \cite{wu_relation-aware_2019}, the sensitivity to different proportions of initial seeding alignments was analyzed, and both models improved with an increased amount of seeding alignments as expected.
%Their evaluation is regarded as incomplete because changing the amount of seeding alignments could possibly result in incompatible distribution of the seeding alignments against the entire KGs.
%For example, the sampled seeding alignments might contain more entity pairs with the same name or more attributes, which could bias the seeding alignments.
%It is therefore necessary to investigate the role of seeding alignments and especially the name bias and attribute bias within.

\subsection{Benchmarks}
\label{sec:benchmarks}

The current benchmarks used to evaluate the embedding-based methods are typically extracted from DBpedia, Wikidata, and YAGO.
They can be divided into two categories.
The first includes those for cross-lingual entity alignment such as DBP15K \cite{sun2017cross} and WK3l60k \cite{chen2018co}, both of which support the alignment between DBpedia entities in English and DBpedia entities in other languages, such as Chinese or French.
These benchmarks usually only support within KG alignment.
The second includes those for cross-KG entity alignment such as DWY15K \cite{guo2018recurrent} and DWY100K \cite{sun2018bootea}, both of which are for the alignment between DBpedia and Wikidata/YAGO.

As discussed in \newcite{sun_benchmarking_2020}, entities in these aforementioned benchmarks have a significant bias in comparison with normal entities in the original KGs; for example, those DBpedia entities in WK3l60k have an average connection degrees of $22.77$ while that of all DBpedia entities is $6.93$.
Thus, these benchmarks are not representative to DBpedia, Wikidata, and YAGO.
To address this issue, \newcite{sun_benchmarking_2020} proposed a new iterative degree-based sampling algorithm to extract new benchmarks for both cross-lingual entity alignment within DBpedia and cross-KG entity alignment between DBpedia and Wikidata/YAGO.
Although the new benchmarks are more representative w.r.t. the graph structure, the entity labels defined by \textit{rdfs:label} are removed, which include important name information, which makes them less representative to real-world alignment contexts.
More importantly, since DBpedia, Wikidata, and YAGO are constructed from the same source Wikipedia, the entities for alignment often have similar names, attributes, or graph structures.
% \revise{This makes these benchmarks very different from real alignment which in contrast usually aims at KGs from different sources for complementing each other.}
These benchmarks are therefore not applicable in the real-world alignment which in contrast, aims at KGs from different sources to complement each other.
To make an industry evaluation, we constructed a new benchmark from two industrial KGs (cf. Section \ref{sec:ib}). 

It is worth noting that Ontology Alignment Evaluation Initiatives\footnote{\url{http://oaei.ontologymatching.org/}} has been organizing a KG track since 2018 \cite{hertling2020the}.
The benchmarks used are those KGs extracted from several different Wikis from Fandom;\footnote{\url{http://www.fandom.com/}} for example, starwars-swg is a benchmark with mappings between two KGs from Star Wars Wiki and Star Wars Galaxies Wiki.
Multiple benchmarks are adopted, but their scales are limited;
for example, $4$ out of $5$ used in 2019 have less than 2K entity mappings.
%, while the remaining one has around 9K entity mappings.
As the two KGs of a benchmark are about two hubs of one concrete topic (such as the movie and the game of Star Wars), the entity name has little ambiguity and becomes a superior indicator for alignment.
%simple label matching can already achieve good performance and often outperforms all the participated systems.
Thus they are not suitable industrial benchmarks for evaluating the embedding-based entity alignment methods.

% \revise{The recent benchmarking study \cite{sun_benchmarking_2020} reviewed 23 state-of-the-art approaches for embedding-based entity alignment and analysed the strength and limitation of each approach on their proposed benchmark dataset, which shed light on the difference between hand-crafted, contrived datasets and real-world datasets. 
% However, their analysis on the datasets was skewed and insufficient. 
% For instance, in one of their benchmark datasets, D-Y-15K, the aligned entities from both Knowledge Graphs have exactly the same name, and arguably a naive string matching approach could achieve comparative performance. }

\section{Data Generation}
\label{sec:dg}
% Introduce the datasets (gold seeding alignments) we developed -- their purpose (why), generation procedure (how), statistics.
% D-W-15K
% What is the industry and application background in making these samples?

\subsection{Industrial Benchmark}
\label{sec:ib}

To evaluate the embedding-based entity alignment methods in an industrial context as discussed above, we first extract a benchmark from two real-world medical KGs for alignment.
One KG is built upon multiple authoritative medical resources, covering fine-grained knowledge about illness, symptoms, medicine, etc.
It is deployed to support applications such as question answering and medical assistants in our company.
However, some of its entities have incomplete information with many important attributes missing, which limits its usability.
We extract around 10K such entities according to the feedback from downstream applications.
% give more concrete details on the applications?
%
They are then aligned with another KG to improve the information completeness.
That KG is extracted from the information boxes of Baidu Baike\footnote{\url{https://baike.baidu.com/}}, the largest Chinese encyclopedia, via NLP techniques (such as NER and RE) as well as some handcrafted engineering work.
We refer to crowdsourcing for annotating the mappings, where heuristic rules, based on labels and synonyms, and a friendly interface for supporting information check are used for assistance.
Finally, we obtain $9,162$ one-to-one entity mappings, based on which one sub-KG is extracted from one original KG.
Specifically, the sub-KG includes triples that are composed of entities associated with these mappings.
The two sub-KGs are named as MED and BBK, and the new benchmark is named as MED-BBK-9K.
% The two sub-KGs are named as MED and BBK, while the new benchmark, composed of MED, BBK, and the $9,162$ one-to-one mappings, is named as MED-BBK-9K.

% \footnote{The benchmark is publicly available at \url{https://github.com/ZihengZZH/industry-eval-EA}.}.

\begin{table}[ht]
    \centering
    \small
    \renewcommand{\arraystretch}{1.1}
    \caption{Statistics of MED-BBK-9K and D-W-15K.}
    \begin{tabular}{p{2.0cm}<{\centering}|p{1.2cm}<{\centering}|p{1.1cm}<{\centering}|p{1.4cm}<{\centering}p{1.3cm}<{\centering}p{0.9cm}<{\centering}|p{1.4cm}<{\centering}p{1.3cm}<{\centering}p{0.9cm}<{\centering}}
    % \begin{tabular}{c|c|c|ccc|ccc}
        \Xhline{2\arrayrulewidth}
        \multirow{2}*{Benchmark} & \multirow{2}*{KGs} & \multirow{2}*{\#Entities} & \multicolumn{3}{c|}{Relation} & \multicolumn{3}{c}{Attribute} \\
        ~ & ~ & ~ & \#Relations & \#Triples & Degree & \#Attributes & \#Triples & Degree \\
        \hline
        \multirow{2}*{MED-BBK-9K} & MED & 9,162 & 32 & 158,357 & 34.04 & 19 & 11,467 & 1.24 \\
        & BBK & 9,162 & 20 & 50,307 & 10.96 & 21 & 44,987 & 4.91 \\
        \hline
        \multirow{2}*{D-W-15K} & DBpedia & 15,000 & 167 & 73,983 & 8.55 & 175 & 66,813 & 4.40 \\
        ~ & Wikidata & 15,000 & 121 & 83,365 & 10.31 & 457 & 175,686 & 11.59 \\
        \Xhline{2\arrayrulewidth}
    \end{tabular}
    \label{tab:dataset_stats}
\end{table}

\begin{figure}[ht]
    \centering
    \includegraphics[width=0.95\linewidth]{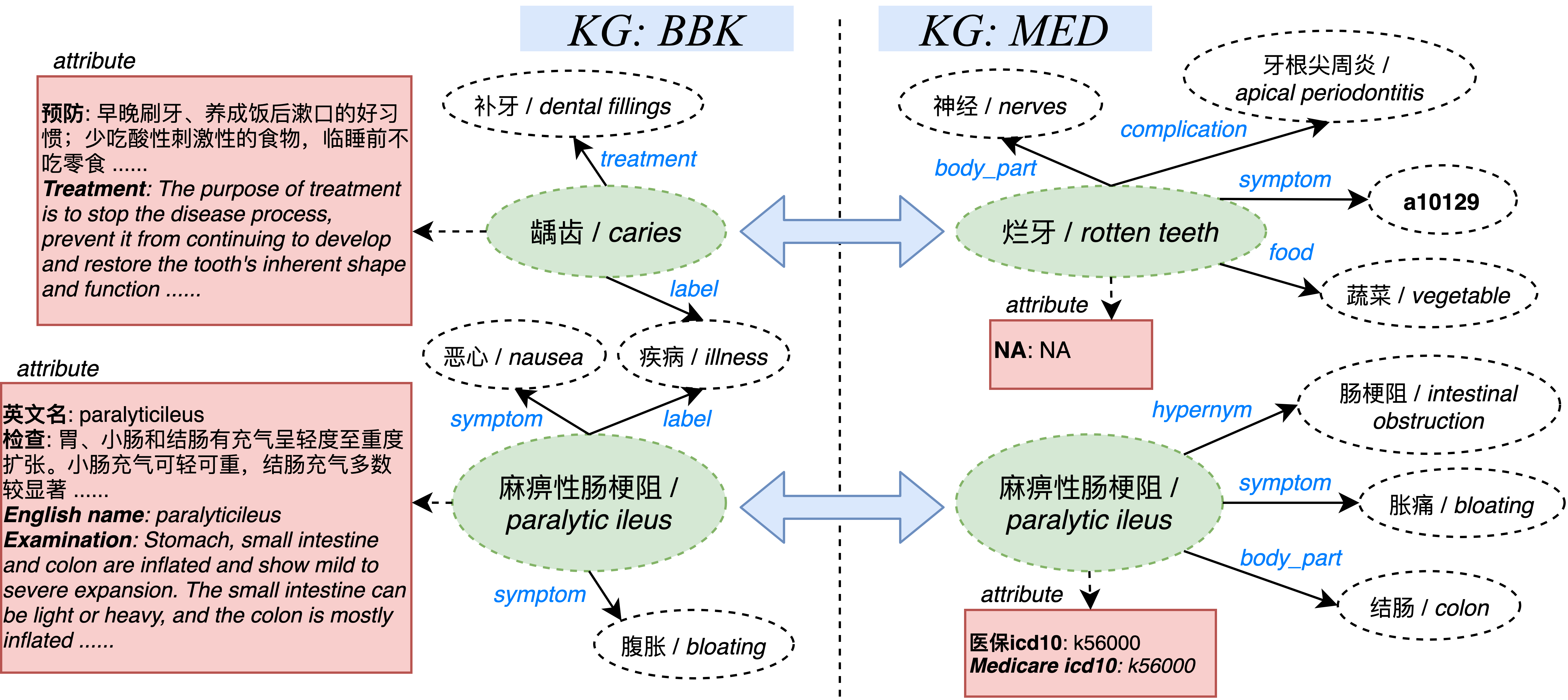}
    \caption{Two mapping examples from MED-BBK-9K with \textit{English translations}.}
    \label{fig:case}
\end{figure}

More details of MED-BBK-9K and another benchmark D-W-15K, which is extracted by the iterative degree-based sampling method under the setting of V2 \cite{sun_benchmarking_2020}, are shown in Table~\ref{tab:dataset_stats}, where \# denotes the number and degree is the rate between the triple number and the entity number.
Statistics of relation triples and attribute triples are separately presented in Table~\ref{tab:dataset_stats}.
Note that a relation is equivalent to an object property connecting two entities, while an attribute is equivalent to a data property associating an entity with a value of some data type.
Two entity mapping examples of MED-BBK-9K are depicted in Figure~\ref{fig:case}, where the green ellipses indicate the aligned entities across KGs, the white ellipses and the solid arrows indicate their relation triples\footnote{\textit{label} here indicates a specific relation. Please do not be confused with \textit{rdfs:label} of the W3C standard.}, and the red rectangles and the dash arrows indicate the attributes which include normal values, sentence descriptions, and noisy values.
Through the statistics and the examples, we can conclude that KGs in MED-BBK-9K are quite different from KGs in D-W-15K, with a higher relation degree, less attributes, higher heterogeneity, etc. 

%
%\#Entities denotes the number of entities, \#Rel./\#Attr. denotes the number of relation/attribute categories, \#Rel tr./\#Attr tr. denotes the number of relation/attribute triples, and Deg. denotes the relation/attribute degree, namely the ratio of \#Rel tr./\#Attr tr. to \#Entities.
%We can find that BBK is much lower in the relation degree but much higher in attribute degree than MED.
%which is often inevitable in real-world scenarios as knowledge is usually split into several different KGs without a universal schema.
%\add{Furthermore, some attributes in both \textit{MED} and \textit{BBK} are descriptive sentences, such as \textit{clinical\_manifestations}, which are dissimilar to the attributes in general-purpose KGs, DBpedia and Wikidata.}

%A certain proportion of entities in \textit{BBK} are expressed in colloquial language while entities in \textit{MED-BBK-9K} are typically formal expression as the authoritative medical resources.
%In the medical domain, we notice that the number of categories for either relations (e.g., \textit{is\_complication\_of}) or attributes (e.g., \textit{national\_pharmaceutical\_standard}) is also limited when compared with more general KGs like DBpedia or Wikidata in D-W-15K benchmark.
%\revise{The detailed distribution of entities in MED-BBK-9K refers to Figure~\ref{fig:heatmap}.}

\subsection{Biased Seed Mappings}

% Besides the industrial benchmark, we also develop a new approach to extract biased seed mappings that match the industrial context.
% To extract the named biased seed mappings, we first calculate the name similarity score of the entities of each mapping, denoted as $s_{name}$ which is in $\left[0,1\right]$.
% Specifically  $s_{name}$ is the normalized Levenshtein Distance -- an edit distance metric \cite{navarro2001guided} of the two name strings.
% We then rank all the mappings from high similarity to low similarity, and select the top $\alpha\%$ (e.g., $6\%$) as seed mappings for training and validating, and the remaining for testing.
% The seed mappings with an attribute bias is calculated in a close way.
% We calculate the average number of attributes of the entities of each mapping, denoted as $n_{attr}$, and rank all the mappings from more attributes to less attributes.

% \revise{Did we do any preprocessing towards the name before calculating the edit distance? How to deal with low case and high case letters?}

% We also extract seed mappings with both name bias and attribute bias.

% \revise{How do we get seed mappings with both name bias and attribute bias?}

Besides the industrial benchmark, we also develop a new approach to extract biased seed mappings for the industrial context.
We first introduce two variables, $s_{name}$ and $n_{attr}$, in which $s_{name}$ is the normalized Levenshtein Distance -- an edit distance metric \cite{navarro2001guided} in $\left[0,1\right]$ for the name strings of entities of each mapping, and $n_{attr}$ is the average number of attributes of entities of each mapping.
For Wikidata entities in D-W-15K, we use the attribute values of \textit{P373} and \textit{P1476} as the entity names, while for DBpedia entities we use the entity name in the URI. 
Note when one or both entities in one mapping has multiple names, we adopt the two names leading to the highest similarity i.e., the lowest $s_{name}$.
Meanwhile, all the names are pre-processed before calculating $s_{name}$: dash, underline and backslash are replaced by the white space, punctuation marks are removed, letters are transformed into lowercase.

With $s_{name}$ and $n_{attr}$ calculated, we divide all the mappings into three different splits according to either the name similarity or the attribute number.
For the name similarity, the mappings are divided into ``same'' ($s_{name}$=$1.0$), ``close'' ($s_{name}$ $<$ $1.0$) and ``different'' ($s_{name}$ is NA, i.e., no valid entity name) for both MED-BBK-9K and D-W-15K.
From the attribute number, the mappings are divided into ``large'' ($n_{attr}\geq k_1$), ``medium'' ($k_2\leq n_{attr}<k_1$) and ``small'' ($n_{attr}<k_2$), where $(k_1,k_2)$ are set to $(5,2)$ for MED-BBK-9K and set to $(10,4)$ for D-W-15K.

%With aforementioned splits, we detail our sampling strategy for biased seed mappings, and we start by presenting how to extract seed mappings with both name bias and attribute bias.
We further develop an iterative algorithm to extract the seed mappings with name bias and attribute bias.
Its steps are shown below, with two inputs, namely the set of all the mappings $\mathcal{M}_{all}$ and the size of seed mappings $N_{seed}$, and one output, namely the set of biased seed mappings $\mathcal{M}_{seed}$.

\setenumerate[1]{label=(\arabic*),parsep=-3pt}
\begin{enumerate}
    \item Initialize the biased seed mapping set $\mathcal{M}_{seed}$.
    \item Assign each mapping in $\mathcal{M}_{all}$ a score: $z = z_{name} + z_{attr}$, where $z_{name}$ is set to $4$, $3$ and $1$ if the mapping belongs to ``same'', ``close'' and ``different'' respectively, and $z_{attr}$ is set to $4$, $3$ and $1$ if the mapping belongs to ``large'', ``medium'' and ``small'' respectively.
    Note all the mappings in $\mathcal{M}_{all}$ are assigned a score of $8$, $7$, $6$, $5$, $4$, or $2$. 
    \item Move the mapping with the highest score in $\mathcal{M}_{all}$ to $\mathcal{M}_{seed}$. Randomly select one if multiple mappings in $\mathcal{M}_{all}$ have the highest score. 
    \item Check whether the size of $\mathcal{M}_{seed}$ has been equal to or larger than $N_{seed}$.
    If yes, return  $\mathcal{M}_{seed}$; otherwise, go to Step (3). %continue to append following other available mappings into $B$, and proceed the iteration.
\end{enumerate}

With the above procedure, we can also obtain seed mappings that are name biased alone by setting $z = z_{name}$, and seed mappings that are attribute biased alone by setting $z = z_{attr}$. 
Note the seed mappings $\mathcal{M}_{seed}$ include both training mappings and validation mappings. In our experiment, the former occupies two thirds of the seed mappings while the latter occupies one third.

% \footnote{The source code is publicly available at \url{https://github.com/ZihengZZH/industry-eval-EA}.}. 

%\section{Experimental Setting}
%\input{sections/04-setting}

\section{Evaluation}

\subsection{Experimental Setting}
%In this section, we detail our experimental settings, the generation of biased seed mappings, and the evaluation metrics from different aspects.

% Furthermore, we reduce the number of available seeds as two different settings, 4\% and 2\% of the dataset as training data with corresponding 2\% and 1\% as the validation data.
We first conduct the overall evaluation (cf. Section \ref{sec:or}).
Specifically, the methods BootEA, MultiKE, RDGCN, and RSN4EA are tested under \textit{(i)} an \textbf{industrial context} where the seed mappings are both name biased and attribute biased, and the rate of training (resp. validation) mappings is $2\%$ (resp. $1\%$), and \textit{(ii)} an \textbf{ideal context} where the seed mappings are randomly selected without bias, and the rate of training (resp. validating) mappings is $20\%$ (resp. $10\%$).
We then conduct ablation studies where three impacts of seed mappings are independently analysed, including size, name bias, and attribute bias.

In both overall evaluation and ablation studies, we calculate metrics Hits@$1$, Hits@$5$, and mean reciprocal rank (MRR) with all the testing mappings.
For each testing mapping, the candidate entities (i.e., all the entities in the target KG) are ranked according to their predicted scores; Hits@$1$ (resp. Hits@$5$) is the ratio of testing mappings whose ground truths are ranked in the top $1$ (resp. $5$) entities; MRR is the Mean Reciprocal Rank of the ground truth entity.
Meanwhile, to further analyse the impact of the seed mappings on different kinds of testing mappings, we divide the testing mappings into two three-fold splits -- ``same'', ``close'' and ``different'' from the name biased aspect, and ``small'', ``medium'' and ``large'' from the attribute biased aspect.

We adopt the implementation of BootEA, MultiKE, RDGCN, and RSN4EA in OpenEA, while their hyperparameters are adjusted with the validation set.
%which takes a third of the seed mappings (another two thirds for training).
Specifically, the batch size is set to $5000$, the early stopping criterion is set to when Hits@$1$ begins to drop on the validation set (checked for every $10$ epochs), the maximum epoch number is set to $2000$. 
As MultiKE and RDGCN utilize literals, the word embeddings are produced using a fastText model pre-trained on Wikipedia 2017, UMBC webbase corpus and statmt.org news dataset\footnote{The word embeddings are publicly available at \url{https://fasttext.cc/docs/en/english-vectors.html}.}.
To run them on MED-BBK-9K, the Chinese word embeddings are obtained via a medical-specific BERT model pre-trained on big medical corpora from Tencent Technology\footnote{Other Chinese word embedding models would suffice to reproduce comparable experimental results.}.

%We conduct the experiments on the four comparative models, including BootEA, MultiKE, RDGCN, and RSN4EA, on the same codebase OpenEA proposed by \newcite{sun_benchmarking_2020} and \revise{all these models are experimented with the default hyperparameters}.

%The experiments are concentrated on the introduction of name bias and attribute bias into the seed mappings, and the evaluation is accordingly split into two aspects: ``overall" evaluation and ``multifaceted" evaluation.
%In the overall evaluation, we use Hits@$1$, Hits@$5$, and MRR as the evaluation metrics for the model performance on the entire test set.
%\revise{The multifaceted evaluation, on the other hand, aims to investigate the performance gap, in terms of Hits@$1$, between biased parts in the test set.
%To be more specific, name biased parts include three parts: ``same" (both entity names being completely same), ``relev." (both entity names being slightly different), and ``diff." (one of entity names being garbled or symbolized), and attribute biased parts include non-overlapped intervals representing the number of attributes for each entity.}
% the gap between which and which?
% the dividing of testing samples into ``same'', ``relev.'' and ``diff.'' need quantified descriptions w.r.t. name and attribute.

We finally compare these embedding-based methods with a state-of-the-art conventional system named PARIS (v0.3)\footnote{\url{http://webdam.inria.fr/paris/}}, which is based on lexical matching and iterative calculation of relation mappings, class mappings and entity mappings with their correlations (logic consistency) considered \cite{suchanek2011paris}. 
We adopt the default hyperparameters to PARIS.
Note that PARIS requires no seed mappings for supervision.
As PARIS does not rank all the candidate entities, we use Precision, Recall, and F1-score as the evaluation metrics.
For the embedding-based methods, Hits@$1$ in our one-to-one mapping evaluation is equivalent to Precision, Recall, and F1-score.

% \revise{Which implementation of PARIS do we use?}

% \add{For fairly comparison, we adopt the default hyper parameters to PARIS.}

% \revise{How do we set the thresholds for both PARIS and those embedding-based methods?}

% \add{We don't set thresholds for embedding-based methods as we calculate metrics like hits@k with candidates. It's not necessary to set a threshold.}

\subsection{Overall Results}
\label{sec:or}

Table~\ref{tab:industry} presents the results of those embedding-based methods on both D-W-15K and MED-BBK-9K under the ideal context and the industrial context. 
On one hand, we find that \textit{the performance of all four methods dramatically decreases when the testing context is moved from the ideal to the industrial}, the latter of which is much more challenging with less and biased seed mappings.
For instance, considering the average MRR of all four methods on all testing mappings, it drops from $0.661$ to $0.262$ on D-W-15K, and from $0.327$ to $0.118$ on MED-BBK-9K.

\begin{table}[ht]
    \centering
    \small
    \renewcommand{\arraystretch}{1.1}
    \caption{Overall results under the ideal context and the industrial context.}
    \begin{tabular}{l|l|p{1.3cm}|p{1.0cm}<{\centering}p{1.0cm}<{\centering}p{1.0cm}<{\centering}|p{1.0cm}<{\centering}p{1.0cm}<{\centering}p{1.0cm}<{\centering}|p{1.0cm}<{\centering}p{1.0cm}<{\centering}p{1.0cm}<{\centering}}
        \Xhline{3\arrayrulewidth}
         \multicolumn{2}{c|}{} & \multirow{2}*{Models} & \multicolumn{3}{c|}{Name-based Splits (Hits@$1$)} & \multicolumn{3}{c|}{Attr-based Splits (Hits@$1$)} & \multicolumn{3}{c}{All Test Mappings} \\
         \multicolumn{2}{c|}{} & & Same & Close & Diff. & Small & Medium & Large & Hits@$1$ &Hits@$5$ & MRR \\
         \Xhline{1.5\arrayrulewidth}
         \parbox[t]{2mm}{\multirow{8}{*}{\rotatebox[origin=c]{90}{D-W-15K}}} &  \parbox[t]{2mm}{\multirow{4}{*}{\rotatebox[origin=c]{90}{Ideal}}} & BootEA & .868 & .902 & .753 & .721 & .821 & .912 & .818 & .922 & .864 \\
         & & MultiKE & .977 & .254 & .216 & .306 & .488 & .661 & .484 & .622 & .554 \\
         & & RDGCN & .942 & .934 & .305 & .330 & .734 & .827 & .629 & .756 & .687 \\
         & & RSN4EA & .718 & .718 & .579 & .536 & .663 & .753 & .650 & .797 & .717 \\
         \cline{2-12}
         & \parbox[t]{2mm}{\multirow{4}{*}{\rotatebox[origin=c]{90}{Industrial}}} & BootEA & .050 & .051 & .023 & .015 & .040 & .053 & .037 & .092 & .065 \\
         & & MultiKE & .968 & .211 & .036 & .086 & .392 & .605 & .368 & .426 & .402 \\
         & & RDGCN & .945 & .872 & .062 & .110 & .559 & .759 & .489 & .539 & .514 \\
         & & RSN4EA & .055 & .060 & .029 & .016 & .046 & .065 & .043 & .092 & .068 \\
         \Xhline{1.5\arrayrulewidth}
         \parbox[t]{2mm}{\multirow{8}{*}{\rotatebox[origin=c]{90}{MED-BBK-9K}}} &  \parbox[t]{2mm}{\multirow{4}{*}{\rotatebox[origin=c]{90}{Ideal}}} & BootEA & .334 & .259 & .328 & .388 & .201 & .265 & .307 & .495 & .399  \\
         & & MultiKE & .342 & .173 & .072 & .269 & .149 & .195 & .213 & .367 & .289 \\
         & & RDGCN & .550 & .217 & .056 & .348 & .270 & .242 & .306 & .425 & .365 \\
         & & RSN4EA & .238 & .121 & .226 & .277 & .114 & .095 & .195 & .311 & .253 \\
         \cline{2-12}
         & \parbox[t]{2mm}{\multirow{4}{*}{\rotatebox[origin=c]{90}{Industrial}}} & BootEA & .006 & .003 & .003 & .006 & .002 & .004 & .004 & .011 & .010 \\
         & & MultiKE & .303 & .149 & .041 & .218 & .137 & .155 & .179 & .322 & .252 \\
         & & RDGCN & .329 & .083 & .013 & .201 & .120 & .086 & .158 & .239 & .199 \\
         & & RSN4EA & .008 & .002 & .007 & .009 & .001 & .000 & .005 & .013 & .011 \\
        \Xhline{3\arrayrulewidth}
    \end{tabular}
    \label{tab:industry}
\end{table}

We also find that the performance decreasement, when moved to the industrial context, varies from one testing mapping split to another.
Considering the name-based splitting, the decreasement is the most significant on the ``different'' split, and the least significant on the ``same'' split.
Take MultiKE on MED-BBK-9K as an example, its Hits@$1$ decreases by $11.4\%$, $13.9\%$ and $43.1\%$  on the ``same'', ``close'' and ``different'' splits respectively.
As a result, the methods including MultiKE and RDGCN perform better on the ``same'' split than on the ``close'' and the ``different'' splits.
It meets our expectations because the seed mappings in the industrial context, which are sampled with a bias toward those with high name similarity, are close to the ``same'' split and far away from the ``different'' split.
However, such a regular is violated when we consider the attribute based seed mapping splits.
As to MultiKE tested by the ``large'' testing split, its performance decreasement when moved to the industrial context is the least significant on D-W-15K, which is as expected, but is the most significant on MED-BBK-9K.
Thus MultiKE performs worse on the ``large'' testing split than on the ``small'' testing split (with $28.9\%$ lower Hits@$1$), although the former is more close to the seed mappings.
One potential explanation is that mappings with more than $5$ attributes (mappings in the ``large'' testing split) in MED-BBK-9K tend to have duplicate attributes and some attribute values are sentences that cannot be fully utilized by these methods.
%The decreasement of both BootEA and RSN4EA is extremely large on all the attribute based splits.

%In this section, we aim to analyse the overall impact introduced by the biased seed mappings w.r.t limited size, name bias, and attribute bias.
%We conduct experiments with the real-world setting on both datasets, and the experimental results for D-W-15K and MED-BBK-9K are presented in Table~\ref{tab:results} and Table~\ref{tab:industry} respectively.

%As shown in Table~\ref{tab:results}, all four models show similar performance in Hits@$1$, Hits@$5$ and MRR in the real-world setting and name-biased or attr-biased settings. For instance, in the tr-2\% setting, MultiKE has Hits@$1$ of 0.360 when being trained on name-biased seeds and Hits@$1$ of 0.384 when being trained on attr-biased seeds, but in the real-world setting, the Hits@$1$ for MultiKE is 0.368, indicating no improvement over solely name-biased or attr-biased. From the multifaceted view, the performance in different splits also show similar patterns as the findings in Section~\ref{name_bias_impact} and \ref{attr_bias_impact}.

On the other hand, we find that \textit{MultiKE and RDGCN are much more robust than BootEA and RSN4EA in the industrial context on both D-W-15K and MED-BBK-9K}.
Although MultiKE and RDGCN do not perform as well as in the ideal context, their performance is still promising. 
Specifically, when measured by all testing mappings, RDGCN performs better than MultiKE on D-W-15K with $27.9\%$ higher MRR and $32.9\%$ higher Hits@$1$ but performs worse than MultiKE on MED-BBK-9K with $21.3\%$ lower MRR and $11.7\%$ lower Hits@$1$.
The performance of BootEA and RSN4EA is poor in the industrial context; their Hits@$1$, Hits@$5$, and MRR on all testing mappings or on different testing splits are all lower than $0.1$ for both benchmarks.
This means that they are very sensitive to the size or/and the bias of the seed mappings (cf. Section \ref{sec:as} for the ablation studies).

%Notice that MultiKE and RDGCN considerably outperform BootEA and RSN4EA in the real-world setting on both D-W-15K and MED-BBK-9K datasets. The overall findings in the experiments on MED-BBK-9K is similar to that on D-W-15K except for three following aspects.
%Firstly, in the testing pairs with relevant sames, MultiKE and RDGCN perform much worse on MED-BBK-9K than on D-W-15K. We suppose that the entity pairs with relevant names in MED-BBK-9K are typically medical terms with colloquial and formal expressions, whose word embeddings are probably not similar.
%Secondly, MultiKE and RDGCN fail to achieve higher Hits@$1$ in the testing pairs with more than 5 attributes in MED-BBK-9K, a clear conflict to the findings in D-W-15K. The accountable reason could be that most attribute-rich entities have duplicate attribute keys or descriptive sentences as attribute values, bringing no contribution to entity embeddings.
%Last, fairly speaking, all four models, especially BootEA and RSN4EA, fail to output alignments in the real-world setting on MED-BBK-9K, which suggests the current embedding-based approaches could be not applicable to the industry datasets.

\subsection{Ablation Studies}
\label{sec:as}

\begin{table}[ht]
    \centering
    \small
    \renewcommand{\arraystretch}{1.1}
    \caption{Results on D-W-15K under different settings (biases and ratios) of the training mappings.}
    \begin{tabular}{l|l|p{1.3cm}|p{1.0cm}<{\centering}p{1.0cm}<{\centering}p{1.0cm}<{\centering}|p{1.0cm}<{\centering}p{1.0cm}<{\centering}p{1.0cm}<{\centering}|p{1.0cm}<{\centering}p{1.0cm}<{\centering}p{1.0cm}<{\centering}}
        \Xhline{3\arrayrulewidth}
         \multicolumn{2}{c|}{\multirow{2}*{Settings}} & \multirow{2}*{Models} & \multicolumn{3}{c|}{Name-based Splits (Hits@$1$)} & \multicolumn{3}{c|}{Attr-based Splits (Hits@$1$)} & \multicolumn{3}{c}{All Test Mappings} \\
         \multicolumn{2}{c|}{} & & Same & Close & Diff. & Small & Medium & Large & Hits@$1$ &Hits@$5$ & MRR \\
         \Xhline{1.5\arrayrulewidth}
         \parbox[t]{3mm}{\multirow{12}{*}{\rotatebox[origin=c]{90}{With No Bias}}} & \parbox[t]{2mm}{\multirow{4}{*}{\rotatebox[origin=c]{90}{20\%}}} & BootEA & .868 & .902 & .753 & .721 & .821 & .912 & .818 & .922 & .864 \\
         & & MultiKE & .977 & .254 & .216 & .306 & .488 & .661 & .484 & .622 & .554 \\
         & & RDGCN & .942 & .934 & .305 & .330 & .734 & .827 & .629 & .756 & .687 \\
         & & RSN4EA & .718 & .718 & .579 & .536 & .663 & .753 & .650 & .797 & .717 \\
         \cline{2-12}
         & \parbox[t]{2mm}{\multirow{4}{*}{\rotatebox[origin=c]{90}{4\%}}} & BootEA & .104 & .087 & .092 & .078 & .085 & .125 & .096 & .206 & .153 \\
         & & MultiKE & .975 & .217 & .088 & .159 & .440 & .647 & .413 & .513 & .467 \\
         & & RDGCN & .898 & .901 & .123 & .163 & .650 & .754 & .521 & .605 & .562 \\
         & & RSN4EA & .105 & .079 & .090 & .071 & .078 & .133 & .093 & .168 & .132 \\
         \cline{2-12}
         & \parbox[t]{2mm}{\multirow{4}{*}{\rotatebox[origin=c]{90}{2\%}}} & BootEA & .024 & .022 & .030 & .028 & .025 & .026 & .026 & .073 & .051 \\
         & & MultiKE & .969 & .224 & .048 & .121 & .428 & .639 & .394 & .463 & .433 \\
         & & RDGCN & .900 & .895 & .107 & .147 & .636 & .761 & .513 & .582 & .547 \\
         & & RSN4EA & .026 & .015 & .031 & .025 & .021 & .034 & .027 & .056 & .044 \\
         \Xhline{1.5\arrayrulewidth}
         \parbox[t]{3mm}{\multirow{12}{*}{\rotatebox[origin=c]{90}{With Name Bias}}} & \parbox[t]{2mm}{\multirow{4}{*}{\rotatebox[origin=c]{90}{20\%}}} & BootEA & .871 & .903 & .535 & .433 & .737 & .931 & .645 & .766 & .702 \\
         & & MultiKE & .978 & .285 & .080 & .085 & .230 & .318 & .185 & .335 & .261 \\
         & & RDGCN & .966 & .924 & .111 & .102 & .521 & .641 & .362 & .441 & .402 \\
         & & RSN4EA & .786 & .800 & .391 & .271 & .631 & .827 & .514 & .656 & .580 \\
         \cline{2-12}
         & \parbox[t]{2mm}{\multirow{4}{*}{\rotatebox[origin=c]{90}{4\%}}} & BootEA & .733 & .817 & .358 & .260 & .633 & .802 & .554 & .642 & .596 \\
         & & MultiKE & .971 & .209 & .053 & .106 & .391 & .609 & .358 & .427 & .398 \\
         & & RDGCN & .956 & .905 & .076 & .128 & .616 & .766 & .491 & .544 & .518 \\
         & & RSN4EA & .198 & .185 & .087 & .051 & .147 & .228 & .138 & .228 & .182 \\
         \cline{2-12}
         & \parbox[t]{2mm}{\multirow{4}{*}{\rotatebox[origin=c]{90}{2\%}}} & BootEA & .031 & .031 & .017 & .013 & .026 & .034 & .024 & .069 & .049 \\
         & & MultiKE & .968 & .195 & .027 & .093 & .389 & .617 & .360 & .404 & .388 \\
         & & RDGCN & .956 & .871 & .056 & .118 & .606 & .766 & .490 & .541 & .516 \\
         & & RSN4EA & .054 & .040 & .027 & .018 & .036 & .062 & .038 & .084 & .062 \\
         \Xhline{1.5\arrayrulewidth}
         \parbox[t]{3mm}{\multirow{12}{*}{\rotatebox[origin=c]{90}{With Attribute Bias}}} & \parbox[t]{2mm}{\multirow{4}{*}{\rotatebox[origin=c]{90}{20\%}}} & BootEA & .789 & .870 & .397 & .365 & .734 & .936 & .565 & .682 & .621 \\
         & & MultiKE & .975 & .358 & .078 & .145 & .488 & .767 & .334 & .459 & .398 \\
         & & RDGCN & .946 & .919 & .109 & .168 & .667 & .885 & .437 & .522 & .479 \\
         & & RSN4EA & .725 & .816 & .309 & .277 & .670 & .834 & .489 & .611 & .546 \\
         \cline{2-12}
         & \parbox[t]{2mm}{\multirow{4}{*}{\rotatebox[origin=c]{90}{4\%}}} & BootEA & .704 & .819 & .337 & .245 & .622 & .800 & .538 & .611 & .574 \\
         & & MultiKE & .972 & .211 & .057 & .115 & .430 & .662 & .383 & .450 & .421 \\
         & & RDGCN & .922 & .908 & .091 & .133 & .630 & .798 & .501 & .557 & .529 \\
         & & RSN4EA & .192 & .213 & .083 & .056 & .156 & .228 & .141 & .232 & .185 \\
         \cline{2-12}
         & \parbox[t]{2mm}{\multirow{4}{*}{\rotatebox[origin=c]{90}{2\%}}} & BootEA & .052 & .051 & .023 & .017 & .039 & .059 & .037 & .094 & .066 \\
         & & MultiKE & .968 & .229 & .041 & .104 & .426 & .651 & .384 & .449 & .421 \\
         & & RDGCN & .915 & .895 & .078 & .122 & .615 & .785 & .497 & .552 & .524 \\
         & & RSN4EA & .068 & .073 & .027 & .018 & .050 & .083 & .049 & .096 & .073 \\
        \Xhline{3\arrayrulewidth}
    \end{tabular}
    \label{tab:results}
\end{table}

\subsubsection{Size Impact}
\label{size_impact}

According to the results in the ``With No Bias'' setting in Table \ref{tab:results}, we can first find that \textit{MultiKE and RDGCN are relatively robust w.r.t. a small training mapping size}.
Considering their Hits@$1$ measured on all the test mappings, it drops slightly from $0.484$ to $0.394$ and from $0.629$ to $0.513$ respectively when the training mapping size is significantly reduced from $20\%$ to $2\%$.
On the ``same'' testing split and the ``large'' testing split, both of which are close to the training mappings, the performance of MultiKE and RDGCN keeps relatively good when trained by $2\%$ of the mappings. 
On the other two splits, which are more biased compared with training mappings, the performance of MultiKE and RDGCN, however, decreases more significantly.

Furthermore, we find that \textit{BootEA and RSN4EA are very sensitive to the training mapping size.}
For example, the MRR of BootEA (resp. RSN4EA) measured by all the test mappings decreases from $0.864$ to $0.153$ to $0.051$ (resp. from $0.717$ to $0.132$ to $0.044$) when the training ratio decreases from $20\%$ to $4\%$ to $2\%$.
The performance of BootEA is beyond our expectation as it is a semi-supervised algorithm designed for a limited number of training samples.
Besides all the testing mappings, their performance decreasement is also quite significant on different testing splits including the ``same'' and the ``large''.

%As shown in the baseline setting in Tale~\ref{tab:results}, MultiKE and RDGCN are robust to the reduction of seed mappings with Hits@$1$, dropping slightly from 0.484 to 0.394 and from 0.629 to 0.513, when the training data decrease from tr-20\% to tr-2\%.
%BootEA and RSN4EA, however, show obvious sensitivity towards the number of available seeds as fewer training data lead to much lower Hits@$1$, especially in tr-4\% setting where BootEA drops from 0.818 to 0.096 and RSN4EA drops from 0.650 to 0.093.

%\revise{From the multifaceted view, we find that BootEA and RSN4EA, both of which do not involve surface information of entities (i.e. names), show no obvious performance gap between different name biased parts or attribute biased parts in the test set.}
%The Hits@$1$ for MultiKE and RDGCN gradually decrease when the entity names become different and the number of attributes decrease. 
%The accountable reason could be that MultiKE favours same-name entities due to the highly similar name embeddings but different name embeddings bring obstacles to the multi-view learning.

\subsubsection{Name Bias Impact}
\label{name_bias_impact}

The name bias impact from the seed mappings can be evaluated by comparing the settings of ``With Name Bias'' and ``With No Bias'' in Table \ref{tab:results}.
With $20\%$ of the mappings for training, MultiKE and RDGCN are more negatively impacted by the name bias than BootEA and RSN4EA; for example, the MRR measured by all the test mappings drops by $52.9\%$ and $41.5\%$ respectively, while that of BootEA and RSN4EA drops only by $18.8\%$ and $19.1\%$ respectively. 

Specifically, considering different testing mapping splits, the negative impact on MultiKE and RDGCN mainly lies in the ``different'' split (e.g., Hits@$1$ of RDGCN drops from $0.305$ to $0.111$), while the impact on the ``same'' and the ``close'' is relatively limited and sometimes even positive.
Mappings in the ``different'' testing split, which have very biased distributions as the training mappings, are sometimes known as long-tail prediction cases,
and the above phenomena indicate their universality and difficulty in an industrial context.
On the other hand, the negative impact of name bias on MultiKE and RDGCN is still much less than the negative impact of the small size on BootEA and RSN4EA.
Thus when impacted by both small size (using $2\%$ of the mappings for training) and name bias, BootEA and RSN4EA perform poorly.
It is also worth noting that RDGCN outperforms other methods by a large margin in the ``close'' split under all the experimental settings; for example, its Hits@$1$ reaches $0.905$ and $0.871$ with $4\%$ and $2\%$ training mappings while that for MultiKE is only $0.209$ and $0.195$ respectively.

%The name-biased setting in Table~\ref{tab:results} illustrates the name bias impact of the seed mappings.

%With the introduction of name bias, a significant performance drop is observed for all four models, in which MultiKE is more negatively influenced (0.299 drop in Hits@$1$) than RDGCN, BootEA and RSN4EA.
%Further reduction of the amount of prior alignments (from tr-4\% to tr-2\%) does not necessarily suggest more severe impact and Hits@$1$ even improves in MultiKE.

%Specifically, as the findings in size impact, the negative impact mainly lies in long-tail pairs with different names and less than 4 attributes.
%MultiKE suffers the negative impact of name bias mainly in terms of long-tail testing pairs, and for example, Hits@$1$ drops from around 0.9 to around 0.2 when comparing pairs with the same name and relevant or different names and less than 4 attributes. 
%As to RDGCN, the negative impact of name bias lies in all the testing pairs but more significant in those long-tail pairs. 
%It is also worth mentioning that in the relevant name testing pairs, RDGCN performs the best among the four models especially in the tr-2\% setting.

\subsubsection{Attribute Bias Impact}
\label{attr_bias_impact}

The attribute bias impact from the seed mappings can be analysed by comparing the settings of ``With Attribute Bias'' and ``With No Bias'' in Table \ref{tab:results}.
When $20\%$ mappings are used for training, its negative impact on all four methods are similar; for example, the MRR of BootEA, MultiKE, RDGCN, and RSN4EA on all testing mappings drops by $28.1\%$, $28.2\%$, $30.3\%$, and $23.8\%$ respectively.
The negative impact is especially significant on the ``small'' testing split as its average attribute number is very different from that of the training mappings.
In contrast, the impact on the ``large'' testing split is even positive for all four methods; for example, when trained by $4\%$ of the mappings, Hits@$1$ of RSN4EA increases from $0.133$ to $0.228$.
Especially, under the attribute bias, reducing the training mappings size has limited impact on MultiKE and RDGCN, and sometimes the impact is even positive that for example, the MRR of MultiKE and RDGCN on all testing mappings increases by $5.8\%$ and $10.4\%$ respectively when the training mapping ratio drops from $20\%$ to $4\%$.

\subsection{Comparison with Conventional System}
% PARIS

This subsection presents the comparison between the embedding-based methods and the conventional system PARIS \cite{suchanek2011paris}, using results in both Table \ref{tab:industry} and Table \ref{tab:conventional}.
Note that Hits@$1$ in Table \ref{tab:industry} is equivalent to Precision, Recall, and F1-Score in our evaluation with all one-to-one mappings.
Although PARIS is an automatic system needing no supervision, it still significantly outperforms all four embedding based methods on both D-W-15K and MED-BBK-9K.
%This observation is consistent with \cite{sun_benchmarking_2020}.
On MED-BBK-9K whose two KGs for alignment are more heterogeneous, the outperformance of PARIS is even more significant; for example, the F1-score of PARIS is $0.493$, while the best of the four embedding based methods is $0.307$ (resp. $0.179$) when trained in the ideal (resp. industrial) context.
One important reason we believe is that these embedding based methods ignore the overall reasoning and the correlation of different mappings, while PARIS utilizes them by an iterative workflow and makes holistic decisions.
Luckily, such reasoning capability and inter-mapping correlations can also be considered in the embedding-based methods, and this indicates an important direction for the future industrial application.

%the embedding-based approaches with conventional approaches, LogMap \cite{jimenez2011logmap} and PARIS \cite{suchanek2011paris}, with respect to their different robustness to entity names and the number of attributes.

% necessary to mention LogMap ??
%We experimented with LogMap but it failed to output alignment results on both D-W-15K and MED-BBK-9K datasets because LogMap heavily depends on the similarity between entity names.
%Therefore, we only present the experimental results of PARIS along with four embedding-based approaches on both datasets in Table~\ref{tab:conventional}.
%In the experiments, the embedding-based approaches are evaluated under the baseline setting while PARIS is trained on the entire dataset in an unsupervised manner.

%It is obvious that PARIS outperforms all the embedding-based approaches on both datasets, especially on MED-BBK-9K.
%Current embedding-based approaches are claimed to focus on learning expressive embedding and ignore the alignment inference \cite{sun_benchmarking_2020}, but xxx (more discussion on PARIS)
%Moreover, PARIS tends to have higher precision than recall (because?).
%The comparison encourages future studies to exhaustively investigate the difference between conventional and embedding-based approaches and to propose novel approaches which take advantage of both.

\begin{table}[h]
    \centering
    \small
    \renewcommand{\arraystretch}{1.1}
    \caption{Results of conventional system PARIS on D-W-15K and MED-BBK-9K.}
    \begin{threeparttable}
    \begin{tabular}{c|l|p{1.0cm}<{\centering}p{1.0cm}<{\centering}p{1.0cm}<{\centering}|p{1.0cm}<{\centering}p{1.0cm}<{\centering}p{1.0cm}<{\centering}|p{2.2cm}<{\centering}}
        \Xhline{3\arrayrulewidth}
          \multirow{2}*{Benchmark} & \multirow{2}*{Metric} & \multicolumn{3}{c|}{Name-based Splits} & \multicolumn{3}{c|}{Attr-based Splits } & \multirow{2}*{All Test Mappings} \\
          & & Same & Close & Diff. & Small & Medium & Large & \\
          \Xhline{1.5\arrayrulewidth}
          \multirow{3}*{D-W-15K} & Precision & .998 & .998 & .900 & .868 & .980 & .999 & .956  \\
           & Recall & .980 & .975 & .707 & .640 & .914 & .987 & .846 \\
           & F1-score & .989 & .986 & .792 & .736 & .946 & .993 & .898 \\
          \cline{2-9}
          \Xhline{1.5\arrayrulewidth}
          \multirow{3}*{MED-BBK-9K} & Precision & .910 & .669 & .778 & .879 & .748 & .757 & .814 \\
           & Recall & .505 & .248 & .258 & .417 & .293 & .314 & .354 \\
           & F1-score & .649 & .362 & .388 & .565 & .422 & .444 & .493 \\
          \cline{2-9}
        \Xhline{3\arrayrulewidth}
    \end{tabular}
    % \begin{tablenotes}
    %   \item[*] The corresponding Precision and Recall are the same as the F1-score.
    % \end{tablenotes}
    \end{threeparttable}
    \label{tab:conventional}
\end{table}

\section{Conclusion and Discussion}

In this study, we evaluate four state-of-the-art embedding-based entity alignment methods in an ideal context and an industrial context.
To build the industrial context, a new benchmark is constructed with two real-world KGs, and the seed mappings are extracted with different sizes, different name and attribute biases.
The performance of all four investigated methods dramatically drops when being evaluated in the industrial context, worse than the traditional system PARIS.
Specifically, MultiKE and RDGCN are sensitive to name and attribute bias but robust to seed mapping size; BootEA and RSN4EA are extremely sensitive to seed mappings size, leading to poor performance in the industrial context.

%In our ablation study, we also investigated the impact of biased seed mappings with respect to size bias, name bias, and attribute bias.
%The further comparison of the embedding-based approaches with conventional approach, PARIS, helped gain insights into how to fully utilize the information.

% Based on these empirical findings, the following pipeline seems plausible in the future industrial applications.
% We firstly apply PARIS with a confidence threshold to generate the mappings from two KGs in an unsupervised manner.
% From all the mappings, we sample non-biased seed mappings that are equally distributed in name-based splits and attribute-based splits.
% The last step is to adopt embedding-based approaches, such as MultiKE and RDGCN, to produce cross-KG alignments using the sampled seed mappings.
% Additionally, current approaches seem not to fully utilize attribute information in the entity embeddings, especially in terms of descriptive long sentences as attribute values.
% How to effectively integrate attribute information into entity embeddings is therefore a promising direction for the future work.

Based on these empirical findings, we recommend to specifically design strategies in crowdsourcing (with tool assistance) to ensure the annotated samples in different name and attribute distributions.
In our industrial context where the seed mappings are limited, adopting MultiKE or RDGCN is demonstrated to be a better choice for cross-KG alignments.
Meanwhile, as mentioned in the evaluation, an ensemble of such embedding based methods with PARIS or LogMap, which considers the correlation between mappings, is also a promising solution for better performance.
Finally, we also plan to develop a robust model that can utilize a complete set of attributes, especially those with values of textual descriptions.

\revise{
% Discussion: possibility for deployment, in which cases they methods can be utilized? In ensemble with PARIS? 

% How to do sampling to deploy them? If we cannot get a lot of samples, adopting MultiKE and RDGCN is a better choice, together with strategies to make the sampling more equally distributed.

% What future work shall we do for the above two aspects?
}

\section*{Acknowledgments}
Jiaoyan Chen's contribution is supported by the AIDA project (Alan Turing Institute), the SIRIUS Centre for Scalable Data Access (Research Council of Norway), Samsung Research UK, Siemens AG, and the EPSRC projects AnaLOG (EP/P025943/1), OASIS (EP/S032347/1) and UK FIRES (EP/S019111/1).
% \section*{Acknowledgements}

% The acknowledgements should go immediately before the references.  Do
% not number the acknowledgements section. Do not include this section
% when submitting your paper for review.

\bibliographystyle{coling}
\bibliography{reference}

\end{document}